\documentclass[sigconf]{acmart}

\usepackage{booktabs} 
\usepackage{times}
\usepackage{latexsym}
\usepackage{amsmath}
\usepackage{amssymb}
\usepackage{mathtools}
\usepackage{multirow}

\setcopyright{acmcopyright}

\copyrightyear{2017} 
\acmYear{2017} 
\setcopyright{acmcopyright}
\acmConference[K-CAP 2017]{K-CAP 2017: Knowledge Capture Conference}{December 4--6, 2017}{Austin, TX, USA}
\acmPrice{15.00}
\acmDOI{10.1145/3148011.3154466}
\acmISBN{978-1-4503-5553-7/17/12}

\begin{document}
\title{Learning Knowledge Graph Embeddings \\ with Type Regularizer}

\author{Bhushan Kotnis and Vivi Nastase}

\affiliation{%
  \institution{Heidelberg University}
  \city{Heidelberg} 
  \state{Germany} 
  \postcode{69120}
}
\email{{kotnis,nastase}@cl.uni-heidelberg.de}


\begin{abstract}
Learning relations based on evidence from knowledge repositories relies on processing the available relation instances. Knowledge repositories are not balanced in terms of relations or entities -- there are relations with less than 10 but also thousands of instances, and entities involved in less than 10 but also thousands of relations. Many relations, however, have clear domain and range, which we hypothesize could help learn a better, more generalizing, model. We include such information in the RESCAL model in the form of a regularization factor added to the loss function that takes into account the types (categories) of the entities that appear as arguments to relations in the knowledge base. Tested on Freebase, a frequently used benchmarking dataset for link/path predicting tasks, we note increased performance compared to the baseline model in terms of mean reciprocal rank and hits@N, N = 1, 3, 10. Furthermore, we discover scenarios that significantly impact the effectiveness of the type regularizer.
\end{abstract}

%
%


\maketitle

{\small
\begin{table*}[]

\centering
\begin{tabular}{l|lll|l}
Source Type & Source & Path or Relation & Target & Target Type \\ \hline \hline
$film$ & $star\_wars\_episode\_IV$ & $produced
\_by$ & $george\_lucas$ & $film\_producer$ \\
$person$ & $alexandre\_dumas$ & $people\_profession$ & $writer$ & $profession$ \\
$academic\_post$ & $professor$ & $profession\_people$ & $albert\_einstein$ & $person$ \\ \hline
\end{tabular}
\caption{\textbf{Entity Type Information}: Examples of source and target entity types from Freebase used in the type regularizer.}
\label{table:cats}
\end{table*}
}

\section{Introduction}
Knowledge -- lexical, world and common-sense -- is crucial for tasks such as automated text comprehension and summarization, question answering, natural language dialogue systems. To make such knowledge available for automatic processing, the most common approach is to provide it as a collection of relation triples -- entities or concepts connected by a relation: e.g., (\textit{concept:city:London}, \textit{relation:country\_capital}, \textit{concept:country:UK}). Globally, such collections can be viewed as knowledge graphs (KGs), for example NELL \cite{Carlson2010T}, Freebase \cite{Bollacker2008} and YAGO \cite{Suchanek2007}. In such graphs, nodes (entities/concepts) may be connected by different types of relations. This results in a multi-graph, i.e. a graph with different types of links where a link type corresponds to a relation type.


KGs are known to be incomplete \cite{Min2013}, i.e., a significant number of relations between entities are missing.  Embedding the knowledge graph in a continuous vector space has been successfully used to address this problem \cite{Nickel2012,Bordes2013,Socher2013}. Such models represent the components of the graph, i.e., the entities and relations, using real valued latent factors that encode the structure of the knowledge graph.  For example the latent factor model should be able to recover \textit{Cologne} from the latent representations of \textit{Moselle} and \textit{river\_flowsThrough\_city}.
Examples include the RESCAL \cite{Nickel2012} tensor factorization model, the TransE model \cite{Bordes2013} and their variations \cite{Lin2015,Nickel2016}. We focus on the RESCAL model, one of the most flexible and widely used models. RESCAL is a bilinear model that represents triples as a pairwise interaction of source and target entity latent factors (embeddings) through a matrix that represents the latent factors of the connecting relation. The entity and relation representations induced can be used to predict additional relations -- edges -- between known entities. Table \ref{table:cats} lists a few examples of entity type information in Freebase. 

\begin{figure}
  \hspace{-1.6cm}\includegraphics[scale=0.37]{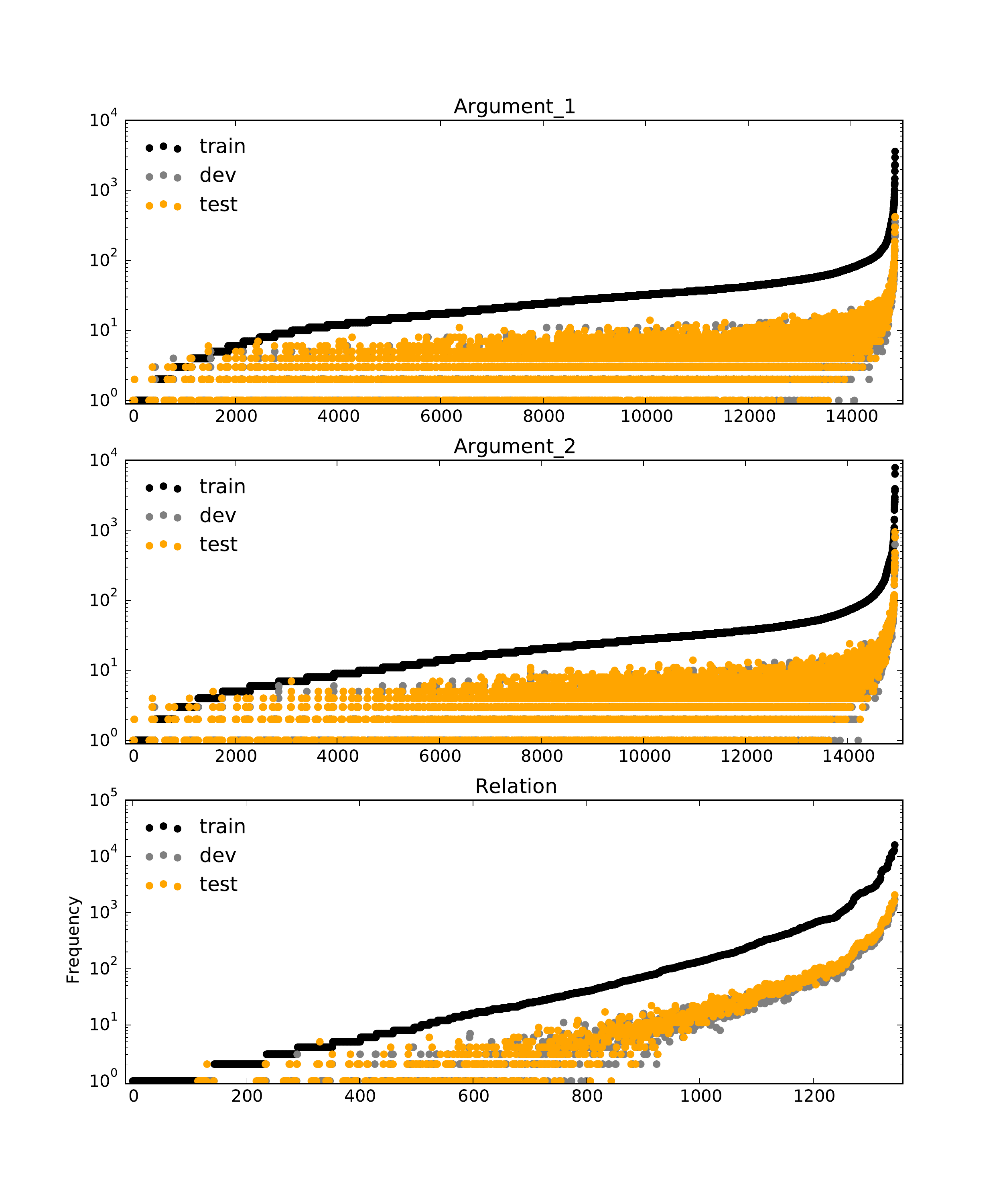}
  \caption{Statistics on argument and relation frequencies for Freebase15k}
  \label{fig:freebasestats}
\end{figure}

Existing knowledge graphs are imbalanced -- both relation and entity frequencies vary widely, as evident from the statistics on Freebase 15k shown in Figure \ref{fig:freebasestats}. Since entity and relation embeddings are based on the connectivity structure of the graph, it is reasonable to ask what is the outcome of the knowledge graph embedding for entities and relations which are underrepresented in the graph, in particular, how good are they for the task of link prediction. 

Approaches such as RESCAL take an extensional view of relations -- they process the collection of instances without knowledge of higher level rules or information about these relations. We hypothesize that providing the higher level -- intensional -- view in the form of types or categories of relation arguments, can lead to improved results for the task of link prediction. This may be true particularly for knowledge graphs such as Freebase that have strongly typed relations, and also for low-frequency relations or for relations involving low-frequency entities.

In this article we present experimental results supporting the hypothesis that augmenting single-relation models with entity type information, in the form of a \textit{`Type'} regularizer,  leads to improvements in predicting missing links. 
The results show that even though the bilinear model induces representations for all entities and relations together -- so it implicitly uses the type information we provide as a separate relation -- the type regularizer which explicitly includes such information for each relation leads to better results. Furthermore, we note the positive impact of including the type regularizer for relations involving low-frequency entities, whereas low-frequency relations are less affected by this added information.
We also analyze the effects of training data size on the usefulness of the type regularizer, and note that its impact grows with the amount of training data. 

\section{Related Work}

A variety of latent factor models \cite{Nickel2012,Bordes2013,Socher2013,Riedel2013} have been developed to represent entities and relations in a knowledge graph, and have been used to address the knowledge base completion (KBC) problem. Most latent factor models are trained on either knowledge graph triples, or triples extracted from open domain knowledge extraction tools \cite{Riedel2013}. A notable exception is the RNN model proposed by \cite{Neelakantan2015} that learns path embeddings for knowledge base completion. \cite{Guu2015} propose a compositional objective function over latent factor models, which is trained on paths as well as triples. For models that are compositional, \cite{Toutanova2016} shows that incorporating intermediate entity information, in the form of latent factors, improves KBC performance. The source and target types are not explicitly included.

\cite{Chang2014} make use the of type information and produce a variation of {\scshape Rescal} they call {\scshape Trescal} -- Typed {\scshape Rescal}. The type information is used to improve the efficiency of the model, by reducing the size of the entity matrix in the computation of the loss function to entities belonging to the domain and range of the relation. The entity type as such is only implicitly incorporated, as something shared by the entities singled out for computing the loss function.

\cite{Das2016} builds on \cite{Neelakantan2015}, and uses an RNN to model paths which incorporate type information for the entities along the path. Entities are represented as a sum of their entity types, which are learned during training. Including this information leads to higher performance. 


Compared with these previous approaches, we add the entity types explicitly in the model, and derive a representation for entities and their types concurrently. We analyze the impact of using such representation for link prediction with different amounts of training data, to understand under what conditions the type information has a positive impact.

\section{Methods}

In this section we describe the RESCAL model and show how the type regularizer was added to include the type information for each relation in the computation of the loss function.

\subsection{Definitions}
Let $\mathcal{E}, \mathcal{R}$ be the set of entities and relations in the KG respectively. A knowledge graph $\mathcal{G}$ is a set of triples $(s,r,t)$ where $s,t \in \mathcal{E}, \ r \in \mathcal{R}$ and relation $r$ connects $s$ to $t$. 

The {\bf knowledge base completion} (KBC) task is the task of 
classifying whether the triple $(s,r,t)$ is a part of the knowledge graph. This can be described as $(s,r,?)$ or $(?,r,t)$ where the question mark represents the unknown correct target/source entity from a set of candidate entities.

\subsection{RESCAL Model}
The RESCAL model \cite{Nickel2012} weights the interaction of all pairwise latent factor between the source and target entity for predicting a relation. It represents every entity as a vector ($x \in \mathbb{R}^d$), and every relation as a matrix $W \in \mathbb{R}^{d\times d}$. This model represents the triple $(s,r,t)$ as a score given by \\
$ sc(s,r,t) = x_s^T \ W_r \ x_t $ \\
This is equivalent to tensor factorization where each relation matrix is a slice of the tensor. These vectors and matrices are learned by constructing a loss function that contrasts the score of a correct triple to incorrect ones. Here we use the max-margin loss  described in the following equation:
\begin{align} 
J(\Theta) = \sum_{i=1}^N \sum_{t' \in N(t)}\textrm{mm}(\sigma(sc_i),\sigma(sc'_i))
\label{eqn:loss}
\end{align}
\vspace{-5mm}
\begin{align*}
 \textrm{mm}(\sigma(sc_i),\sigma(sc'_i)) = max\bigg[0,1 - \sigma(sc_i) + \sigma(sc'_i)\bigg]
\end{align*}
where there are N positive instances, positive and negative instances are scored as $sc_i = sc(s_i,r_i,t_i)$ and $sc^{'}_i = sc(s_i,r_i,t^{'}_i)$, respectively. $N(t)$ is the set of incorrect targets and $\sigma$ is the sigmoid function.

\subsection{The Type Regularizer}
We introduce a regularizer term which incorporates type information of source and target entities. Let $s_{cat}$ be the type for entity $s$ and $r_{cat}$ the relation between $s$ and $s_{cat}$. Depending on the knowledge resource, $r_{cat}$ could be $is\_a$ (in an ontology, for example), $category$ (in a resource built based on Wikipedia), or other such relations that capture the entity type. A few examples of entity types can be seen in Table \ref{table:cats}. Note that entity type information is not used during test time.

If $s$ is the source entity and $t$ the target entity for query $q$, then we define the regularizer as in equation \ref{eqn:tr}, where $N(s_{cat})$ and $N(t_{cat})$ are sets of (negatives) for $s_{cat}, t_{cat}$, while $T(s_{cat}),T(t_{cat})$ are sets of correct categories for source $s$ and target $t$ respectively. $mm$ is the max margin loss described in equation (\ref{eqn:loss}). \\

\vspace{0.4cm}
$R(\Theta,q) :=$
\vspace{-0.2cm}
{
\begin{align}
\label{eqn:tr}
&  \sum\limits_{\substack{s^{'}_{cat} \in N(s_{cat}) \\ s_{cat} \in T(s_{cat})}} \hspace{-0.15in} \textrm{mm} \bigg(\sigma_{sc}(s,r_{cat},s_{cat}), \sigma_{sc}(s,r_{cat},s^{'}_{cat})\bigg) \nonumber \\
&+  \sum\limits_{\substack{t^{'}_{cat} \in N(t_{cat}) \\ t_{cat} \in T(t_{cat})}} \hspace{-0.15in} \textrm{mm} \bigg(\sigma_{sc}(t,r_{cat},t_{cat}), \sigma_{sc}(t,r_{cat},t^{'}_{cat})\bigg) 
\end{align}
}

The complete objective function to be minimized is
\par
\vspace{-0.5cm}
{\small
\begin{align*}
J(\Theta) = \sum_{i=1}^{N} \sum_{t_i^{'} \in N(q_i)}\textrm{mm}(q_i,t_i,t_i^{'}) + \alpha R(\Theta,q_i)
\end{align*}
}
where the hyper-parameter $\alpha$, $\alpha \geq 0$, controls the impact of the regularizer terms and $N(q_i)$ is the set of negative targets for query $q_i$, where $q_i$ corresponds to query $(s_i,r_i,?)$. 

\section{Experiments}

\subsection{Data}
We carry out experiments on FB15K, a subset of the Freebase knowledge graph provided by \cite{Bordes2013}. This dataset is a standard benchmark dataset used for evaluating link prediction algorithms \cite{Bordes2013,Nickel2016,Trouillon2017}. The FB15K dataset consists of 1345 relations and 14,951 entities. The training, validation and test set consists of 483,142, 50,000 and 59,071 triples respectively. The Freebase relations do not include the category relation, thus there is no overlap between the category triples and FB15K triples. 

 We obtain Freebase category data from \cite{Gardner2015}, and then the entity type by mapping the Freebase entity identifier to the Freebase category. This results in 101,353 instances of the {\it category} relation which is used in the training stage. It is not used during test time.

\subsection{Implementation}
We use the Adam \cite{Kingma2014} SGD optimizer for training because it addresses the problem of decreasing learning rate in AdaGrad. We use median gradient clipping to prevent explosive gradients and we also ensure that entity embeddings have unit norm. We performed exhaustive grid search for the L2 regularizer as well as  $\alpha$ on the validation set and we tuned the training duration using early stopping. We use 100 dimensional entity vector in all experiments \footnote{Code is available at https://github.com/bhushank/kge}.

\subsection{Evaluation Procedure}
For evaluation we follow the procedure described in \cite{Socher2013}. For every test triple we predict either the source or the target, and negative instaces for training and testing are produced by corrupting positive ones: we replace $s$ (or $t$) in a $(s, r, t)$ triple with an $s_n$ (or $t_n$) that has the same type as $s$ (or $t$) but does not appear in a positive instance $(s_n, r, t)$ (or $(s, r, t_n)$). For meaningful comparison, the negative triples that occur in training or validation datasets as positive triples are filtered out. For faster evaluation, instead of using all negative triples, we produce 1000 by randomly sampling entities from the entire set. We report results in terms of hits at 1,3,10 (HITS@1,3,10) and mean reciprocal rank (MRR) metrics. Hits at $K$ is the proportion of correct answers (hits) in the first $K$ ranked predictions, while MRR is the mean of the reciprocal of the rank of the correct answers.

\subsection{Results}

We use the bilinear (RESCAL) model as a baseline. As evidenced by the results in Table 2, adding the type regularizer improves performance. It may be tempting to think that the performance improvement is natural since we are providing additional information through the type regularizer. We test this in further experiments.

\begin{table}[h]
\centering
\label{table:comparison}
\begin{tabular}{l|l|l}
  Metrics  & Bilinear & Bilinear + TR   \\
        \hline
MRR     & 0.343    & \textbf{0.3862} \\
HITS@1  & 0.2451         &    \textbf{0.304}             \\
HITS@3  & 0.3804         &    \textbf{0.4161}             \\
HITS@10 & 0.5312   & \textbf{0.5408} \\ \hline
\end{tabular}
\caption{Evaluation: Performance Comparison between bilinear model with and without type regularizer.}
\vspace{-0.7cm}
\end{table}

\begin{table}[]
\centering
\label{table:datasize}
\begin{tabular}{c|c|c|c}
\% training data& Model         & MRR    & \% Improvement \\
\hline
\multirow{2}{*}{100}     & Bilinear      & 0.343  &                        \\
                         & Bilinear + TR & 0.3862 & +12.59                  \\
                         \hline
\multirow{2}{*}{75}      & Bilinear      & 0.3495 &                        \\
                         & Bilinear+ TR  & 0.3552 & +1.6                    \\
                         \hline
\multirow{2}{*}{50}      & Bilinear      & 0.3457 &                        \\
                         & Bilinear + TR & 0.3409 & -1.3                   \\
                         \hline
\multirow{2}{*}{25}      & Bilinear      & 0.332  &                        \\
                         & Bilinear + TR & 0.3198 & -3.67  \\ \hline             
\end{tabular}
\caption{Effect of training data size on TR: Performance comparison between bilinear models with and without type regularizer for different dataset sizes. }
\end{table}

We test the impact of the type regularizer by analyzing its performance on different sizes of training data. We first generate multiple training datasets by randomly sampling 25\%, 50\% and 75\% of the triples. 
As illustrated in Table 3, when using only 25\% to 50\% of the training data, the performance drops. The type regularizer uses category information, under certain circumstances ($\alpha=1$) adding it is equivalent to adding approximately 100,000 new triples with category relation to the training set.
Thus, simply augmenting the model with additional information does not always improve performance.

The reason behind the performance drop with less training data is not obvious, because adding external information should help the model learn better embeddings. We hypothesize that the drop in performance is because when fewer number of training instances are available, the type regularizer leads the system to learn relations that over-generalize. 
The model is biased towards learning categories very well for reducing training loss. This results in embeddings that are biased towards predicting relations at the level of categories and not individual relations resulting in performance drop for the relation prediction task. 

\begin{figure}[h]
\centering
N\includegraphics[scale=0.36,trim={0 0 0 3cm},clip]{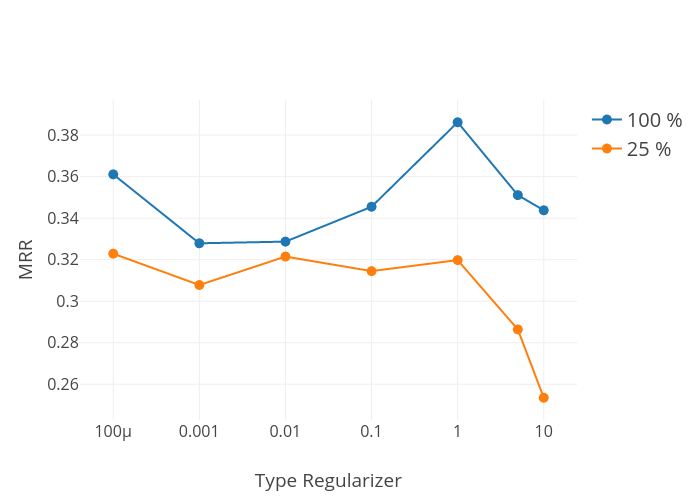}
\caption{MRR vs. $\mathbf{\alpha}$: MRR drops with increasing strength of the type regularizer for models trained on 25\% (blue) and 100\% (orange) of FB15K dataset. Plot for $\mathbf{\alpha = 0.0001,0.001,0.01,0.01,1,5,10}$}
\label{fig:alpha}
\end{figure}

We investigate this hypothesis by varying the value of $\alpha$ that weighs the importance of the type regularizer (cf. equation \ref{eqn:loss}). We plot the Mean Reciprocal Rank vs. the strength of the type regularizer for model trained on only 25\% of the training data in Fig. \ref{fig:alpha}. The higher the strength of the type regularizer, the higher the cost incurred for mis-predicting the category. As Fig. \ref{fig:alpha} shows, MRR falls sharply with increase in $\alpha$. This effect is not observed in the 100\% training data scenario. This suggests that adding category information may lead to improved performance only when the added information does not severely bias the training data.

\begin{table}[h!]
\centering
\begin{tabular}{l|l|l|l}
 & Relation Name             & Instances (train) & Instances (test) \\
\hline
$r_1$    & /people/person/profession & 11636              & 1384           \\
$r_2$    & /music/genre/artists      & 5952               & 679            \\
$r_3$    & /film/film/country        & 2407               & 280            \\
$r_4$    & /tv/tv\_program/genre     & 1010               & 100    \\ \hline
\end{tabular}
\caption{Relations with train and test instances}
\label{tab:rels}
\vspace{-0.7cm}
\end{table}

To investigate the impact of training data size on the type regularizer performance, we analyze in detail the performance of the system for relations with a different number of training instances. Table  \ref{tab:rels} lists four relations we used to look into this phenomenon. 

\begin{figure}
\label{fig:rels}
\centering
\includegraphics[scale=0.36,trim={0 0 0 2cm},clip]{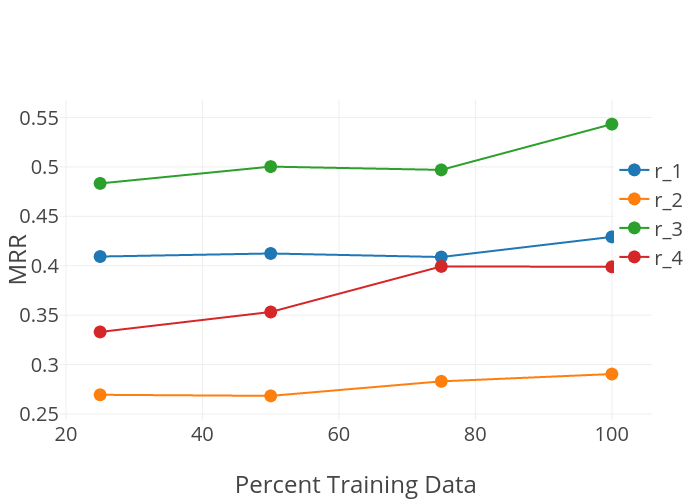}
\caption{MRR vs. Percent Train Data for multiple relations: Number of training instances modulate effect of Type Regularizer. Relations listed in Table \ref{tab:rels} }
\label{fig:nrinstances}
\end{figure}

\begin{figure*}
  \includegraphics[scale=0.4]{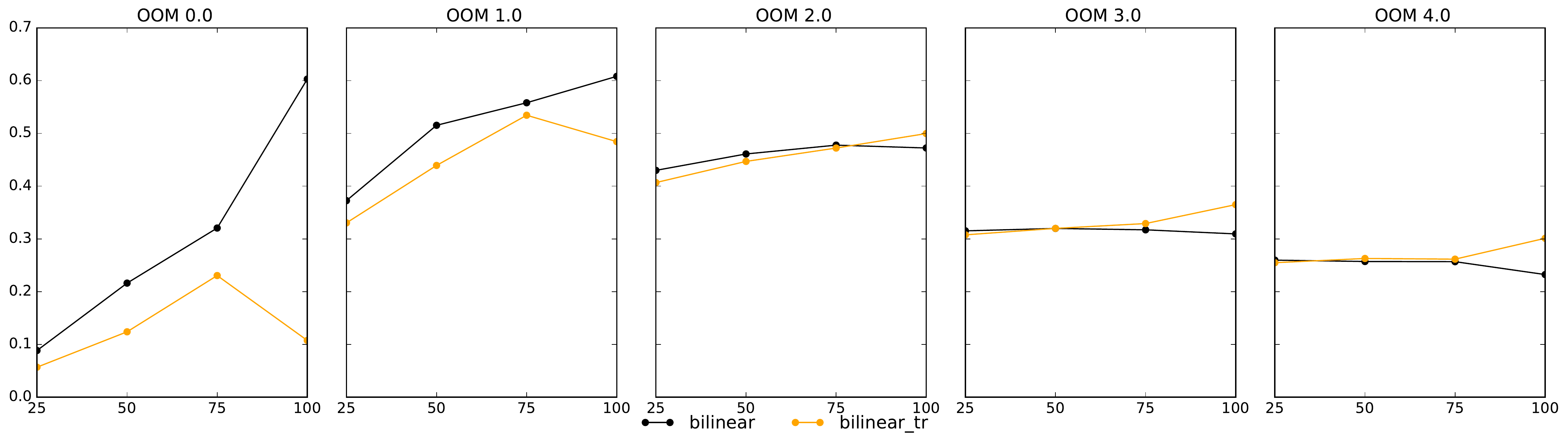} \\
  \includegraphics[scale=0.4]{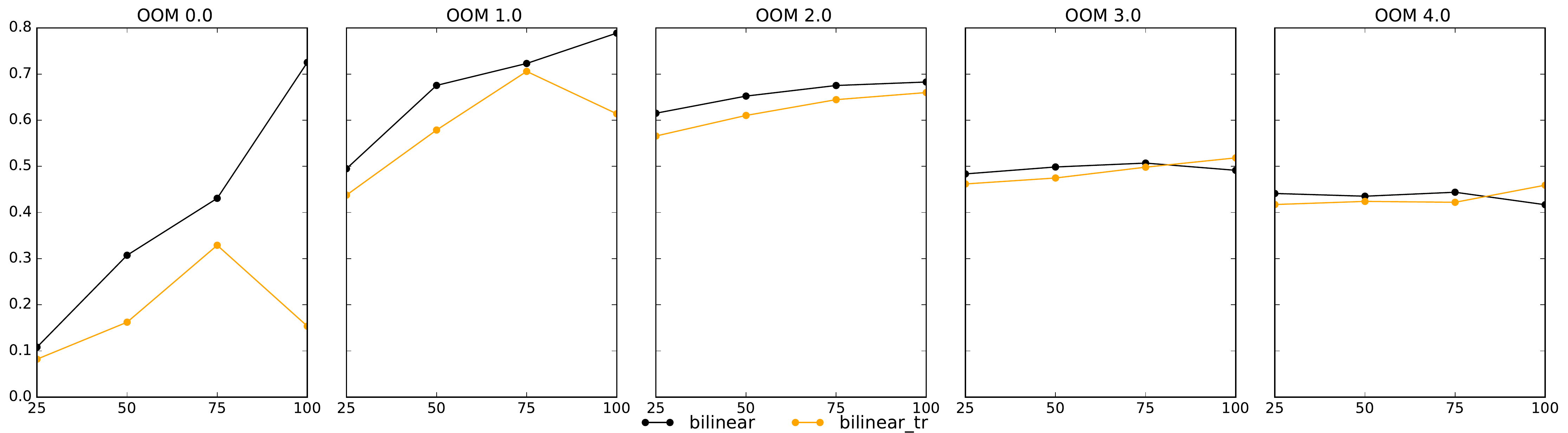}
\caption{MRR and HITS@10 link prediction results grouped by the order of magnitude of relation frequency, for different amounts of training data.}
  \label{fig:rels}
\end{figure*}

\begin{figure*}
  \includegraphics[scale=0.4]{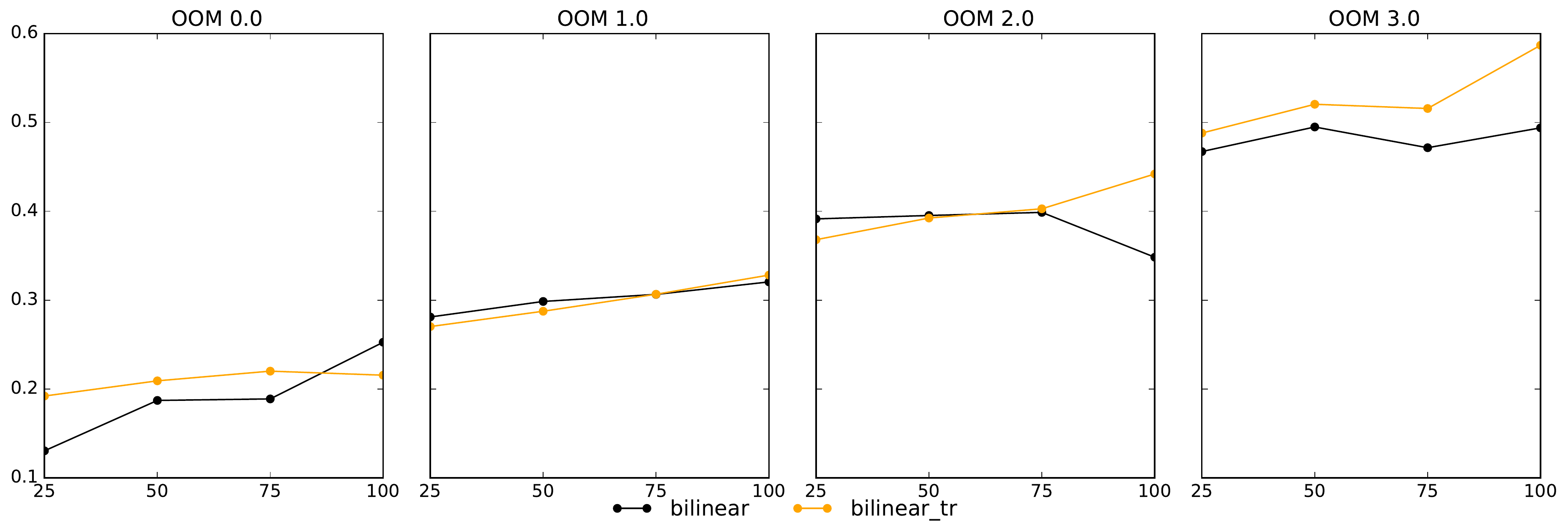} \\
  \includegraphics[scale=0.4]{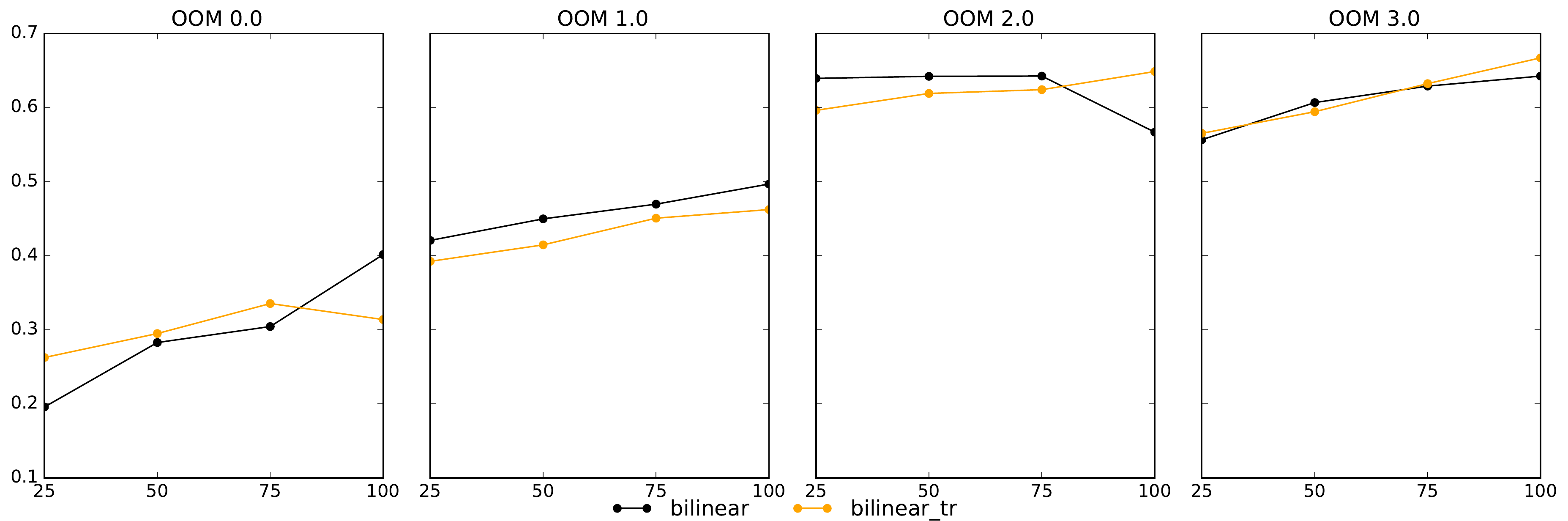}
\caption{MRR and HITS@10 link prediction results grouped by the order of magnitude of entity frequency, for different amounts of training data.}
  \label{fig:args}
\end{figure*}

Fig. \ref{fig:nrinstances} shows the performance in terms of MRR (using Type Regularizer) for link prediction on these four relations. The orange and blue lines denote relations ($r_1,r_2$) with 11,636 and 5952 training instances respectively, while the red and green curves denote relations ($r_3,r_4$) with 2407 and 1010 training instances respectively. The red and green curves (the relations with fewer instances) show a larger change in MRR compared to the orange and blue curves. This confirms our hypothesis that the Type Regularizer is more sensitive for relations with a smaller number of training instances, and indicates that the embeddings learned for relations with larger number of instances are less biased towards predicting categories.   

We note that equation (\ref{eqn:tr}) has the same max margin structure as the loss function, equation (\ref{eqn:loss}). Therefore using this particular formula for the type regularizer is equivalent to adding the {\it category} relation as an additional slice of the tensor factorized by RESCAL, then the hyperparameter $\alpha$ is 1. Experiments have shown though that fine tuning $\alpha$ -- and this fine-tuning the usage of type information -- can lead to better results. More specifically it is equivalent to adding 101,353 unique instances of {\it category} relation.

We also performed overall relation and entity analysis based on their occurrence frequency. Looking at relations grouped by the order of magnitude (oom) of their occurrence frequency presented in Figure \ref{fig:rels} we note that low frequency relations seem not to be affected by the type regularizer, and are modeled better using only the instances themselves. The reson for this is that very low frequency relations actually connect high frequency entities, e.g. relation {\it /award/hall\_of\_fame/discipline}. On the other hand, high frequency relations have overall lower results than other relations. The reason for this is that in numerous cases, one of the arguments of these relations is a low frequency entity. For example, the {\it lives\_in} relation that connects a person with the city they live in, has as the "City" argument an entity that does not appear in many other relations.

To further clarify the reasons for variation in performance of relations, we analyze the link prediction results based on the order of magnitude of entity frequency, presented in Figure \ref{fig:args}. The results in this case are more in line with the expected outcome -- links that involve lower frequency entities have lower prediction results. The type information generally has a positive impact throughout, except medium-range entities where it seems that type information leads to over generalization.

Using the type regularizer as an additional terms whose weight can be calibrated using the $\alpha$ parameter makes it easier to adjust the influence of the type information based on node degrees and relation frequencies. Furthermore, by incorporating the type information in the loss function for every relation as opposed to having it as a separate relation in the knowledge graph allows the incorporation of the range and domain information for each relation, as opposed to modelling the entity type outside of a particular environment.

It is interesting to note that the best results for medium to high frequency entities and relations are obtained when using the full training data and the type regularizer. This indicates that the type regularizer can mitigate the overfitting tendency of RESCAL, and produce a more robust model.

\section{Conclusion}
We proposed a type regularizer that leverages entity type information for state-of-the-art latent factor models like RESCAL. Experiments on Freebase FB15K dataset suggest that adding the type regularizer improves performance on the knowledge base completion task. However adding category information may not improve results for all relations, particularly those with fewer positive instances where introducing category information may lead to embeddings that are biased towards capturing/predicting categories rather than fine grained instances. We plan to study the impact of the added type information for datasets where the relations are not as strongly typed as Freebase -- for grammatical collocation information for example and inducing selectional preferences -- and for more complex, path prediction, tasks.

\bibliographystyle{ACM-Reference-Format}
\bibliography{acl2017} 

\end{document}